\documentclass[sigconf,acmsmall,screen,authorversion]{acmart}
\usepackage{xcolor}
\usepackage{subcaption}
\usepackage{graphicx}

\begin{document}
\setcopyright{acmlicensed}
\acmJournal{PACMCGIT}
\acmYear{2025} \acmVolume{8} \acmNumber{1} \acmArticle{13} \acmMonth{5}\acmDOI{10.1145/3728309}
\title{Non-uniform Point Cloud Upsampling via Local Manifold Distribution}


\author{Yaohui Fang}
\email{202321081054@mail.bnu.edu.cn}
\orcid{0009-0009-0368-2696}
\affiliation{%
    \institution{Beijing Normal University}
    \city{Beijing}
    \country{China}
}

\author{Xingce Wang}
\email{wangxingce@bnu.edu.cn}
\orcid{0000-0002-3177-8902}
\authornote{Corresponding author}
\affiliation{%
    \institution{Beijing Normal University}
    \city{Beijing}
    \country{China}
}


\begin{abstract}
Existing learning-based point cloud upsampling methods often overlook the intrinsic data distribution characteristics of point clouds, leading to suboptimal results when handling sparse and non-uniform point clouds. We propose a novel approach to point cloud upsampling by imposing constraints from the perspective of manifold distributions. Leveraging the strong fitting capability of Gaussian functions, our method employs a network to iteratively optimize Gaussian components and their weights, accurately representing local manifolds. By utilizing the probabilistic distribution properties of Gaussian functions, we construct a unified statistical manifold to impose distribution constraints on the point cloud. Experimental results on multiple datasets demonstrate that our method generates higher-quality and more uniformly distributed dense point clouds when processing sparse and non-uniform inputs, outperforming state-of-the-art point cloud upsampling techniques.
\end{abstract}




\maketitle

\section{Introduction}

Point clouds are a specialized representation of objects in three-dimensional space. In recent years, significant advancements have been made in 3D sensing technologies, leading to their widespread use as input for various 3D applications, such as autonomous driving \cite{wang2019pseudo, lang2019pointpillars}, 3D city reconstruction \cite{lafarge2012creating, musialski2013survey}, and virtual/augmented reality \cite{held20123d, santana2017multimodal}. However, due to practical limitations, raw point clouds generated through 3D scanning are typically noisy, sparse, and unevenly distributed. To ensure the smooth execution of subsequent tasks involving point clouds, it is crucial to perform upsampling to obtain dense, complete, and clean point clouds, while preserving their smoothness and accurately recovering geometric structures.

The goal of point cloud upsampling extends beyond simply generating dense point sets from sparse inputs. More importantly, the generated points should closely approximate the underlying surface and be evenly distributed. Early optimization-based point cloud upsampling methods \cite{alexa2003computing, lipman2007parameterization, huang2009consolidation, huang2013edge} employed various shape priors to constrain point cloud generation, which worked well for simple, smooth surfaces. However, these methods exhibit poor robustness when dealing with complex structures in point clouds. With the rise of deep learning, many deep learning-based methods \cite{yu2018pu, li2019pu, qian2021pu, qian2020pugeo} have been applied to point cloud upsampling tasks, effectively extracting features. However, when dealing with highly sparse and non-uniform point clouds, these networks are unable to effectively learn the geometric structure of the point clouds, resulting in the loss of fine details or structural inaccuracies.

For point cloud upsampling tasks, existing methods typically generate new points by extracting features based solely on spatial positions, without considering the distribution characteristics of the point cloud itself. This limitation results in non-uniform upsampling when applied to sparse and non-uniform point clouds. In this paper, we propose a method that constructs a unified statistical manifold by Gaussian representation for each local neighborhood, and enforces distribution constraints on the manifold to achieve uniform point cloud upsampling.

The contributions of this paper are summarized as follows:
\begin{itemize}
\item We propose a novel point cloud representation that interprets the point cloud as samples drawn from a probability distribution function, establishing a statistical manifold framework for point clouds. The distribution characteristics reflect the intrinsic patterns of data generation. By constructing a statistical manifold, we can better simulate and understand this generative process, as well as model the distribution properties of point clouds, such as point density, clustering regions, and directionalities. These distributional details capture the underlying laws and characteristics of the point cloud data, which are crucial for guiding the upsampling process.
\item We use Gaussian functions to fit the local neighborhoods of the point cloud, where each point on the manifold represents a probability distribution. The overall structure and geometric shape of the point cloud can be described using a statistical model, with the geometric information of the point cloud encoded within the Gaussian functions, thereby providing accurate local geometric guidance for subsequent upsampling.
\item We minimize the geodesic distance between corresponding points on the statistical manifold to constrain the distribution of the point cloud, ensuring that the upsampled point cloud better recovers and preserves the details and shapes of the original data. This approach maintains the consistency of its shape and details, generating a more uniform point distribution.
\end{itemize}

\section{Releated Work}

\subsection{ Optimization-based Point Upsampling}

Optimization-based methods typically require geometric prior information of the point cloud,
such as edges and normals, and their performance tends to degrade for complex shapes. Alexa et al. \cite{alexa2003computing} first introduced a point cloud upsampling algorithm, which adds new points at the vertices of the Voronoi diagram in the local tangent space to achieve point cloud upsampling. Subsequently, Lipman et al. \cite{lipman2007parameterization} proposed a non-parametric method based on the Local Optimal Projection (LOP) operator, which resamples points based on the L1 norm. While this method is robust to noise and outliers, its performance on complex geometries is relatively poor. Later, Huang et al. \cite{huang2009consolidation} developed an improved weighted LOP, which performs upsampling in an edge-aware manner, reducing the quality degradation at sharp edges and corners. Huang et al. \cite{huang2013edge} also proposed an Edge-Aware Resampling (EAR) method, which preserves edges during point cloud upsampling but heavily depends on the provided normal information and parameter tuning. Overall, these methods heavily rely on prior information and place significant demands on the original point cloud.

\subsection{Deep learning-based Point Upsampling}

With the development of deep learning techniques and inspired by the success of PointNet \cite{qi2017pointnet}, many deep learning methods have been proposed for various point cloud processing tasks such as classification, completion, denoising, and other applications. For point cloud upsampling, PU-Net \cite{yu2018pu} was the first to apply neural networks to point cloud upsampling, using PointNet++ \cite{qi2017pointnet++} to extract point features and extending those features through a multi-branch MLP to generate upsampled point clouds. However, this approach does not consider the spatial relationships between points, and thus cannot guarantee the generation of uniformly distributed points. EC-Net \cite{yu2018ec} employs a joint loss of point-edge distances, allowing for the generation of sharp new points at edges. MPU \cite{yifan2019patch} is a progressive method for upsampling point patches, but it struggles to generate new points in missing regions for non-uniform point clouds. PU-GAN \cite{li2019pu} was the first to apply Generative Adversarial Networks (GANs) to point cloud upsampling, introducing uniformity loss and becoming the first method capable of generating uniformly distributed upsampled point clouds. PU-GCN \cite{qian2021pu} uses graph convolutional networks for feature extraction and expansion, proposing a novel point cloud upsampling module called NodeShuffle. PUGeo \cite{qian2020pugeo} is the first method to utilize geometric approaches for point cloud upsampling, improving upsampling results by employing local differential geometry constraints. Dis-PU \cite{li2021point} defines two cascaded subnetworks that complete point cloud upsampling in stages. NP \cite{feng2022neural} introduces the use of neural networks to continuously represent geometric shapes, enabling sampling at arbitrary resolutions. Recently, GeoUDF \cite{ren2023geoudf} introduced a local geometric representation that approximates local shapes using quadratic surfaces. RepKPU \cite{rong2024repkpu} proposed a new paradigm, kernel-to-displacement generation, for point generation. However, existing learning-based methods often fail to achieve good performance when handling highly sparse and non-uniform datasets. Drawing inspiration from LGSur-Net \cite{xiao2024lgsur}, our approach performs Gaussian fitting on the local patches of point clouds and applies further constraints on the statistical manifold constructed from the parameters. Our method generates more uniform results for sparse and non-uniform point clouds.

\begin{figure}[h]
  \centering
  \includegraphics[width=\linewidth]{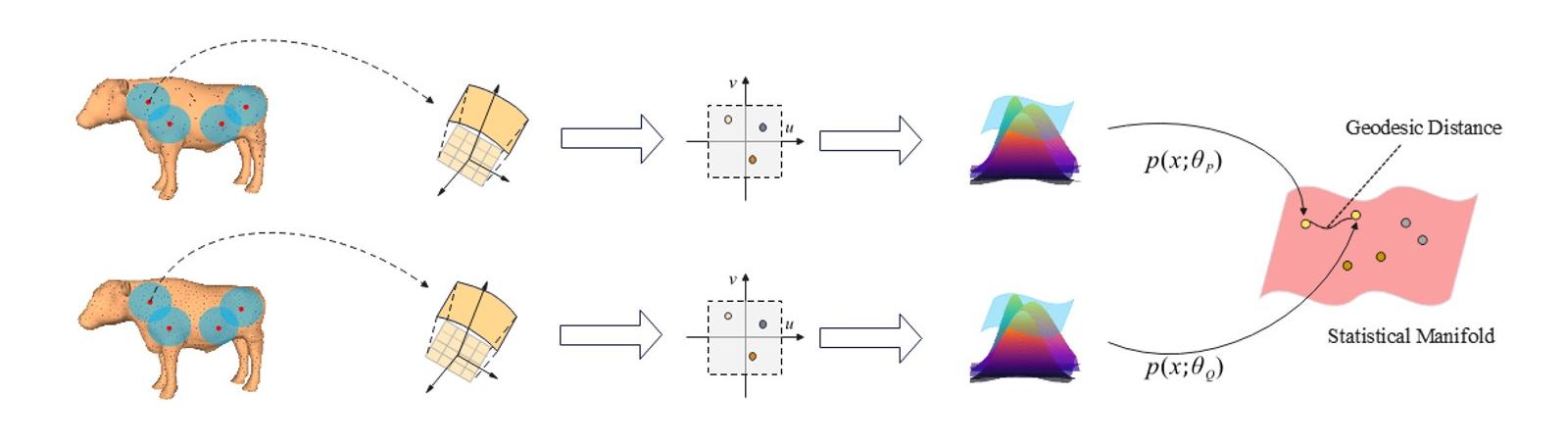}
  \caption{Our upsampling algorithm pipeline. For the input point cloud, we first construct a local coordinate system for each query point based on its surrounding neighborhood. We then perform local Gaussian fitting in the 2D parameter plane. Subsequently, geodesic constraints are applied on the statistical manifold formed by the local Gaussian functions. Finally, by resampling the constructed local Gaussian functions, we generate the upsampled point cloud.}
\end{figure}

\section{Method}

Given a sparse and non-uniform point cloud $P=\left\{p_{i} \in \mathbb{R}^{3 \times 1}\right\}_{i=1}^{N}$ which contains N points, traditional point cloud representations only capture a single 3D position per point, which fails to adequately preserve the underlying distribution characteristics when generating new points. For a given upsampling factor $r$, our goal is to obtain a dense and uniformly distributed upsampled point cloud $P_{r}=\left\{p_{i}|p_{i}^{r}\in \mathbb{R}^{3}\right\}_{i=1}^{rN}$ by starting from each local manifold of the point cloud. Specifically, we perform the following steps. First, a certain number of query points are selected from the point cloud, and for each point $p_{i}$, an overlapping local neighborhood is constructed to define a local coordinate system, an attention network is then employed to generate Gaussian parameters and combination weights, thereby enabling local Gaussian fitting for $p_{i}$. Next, resampling is performed on the projection plane to generate the required number of upsampled points, which are mapped back to the original point cloud through the local Gaussian representation. Finally, each local Gaussian function is integrated into a unified manifold, with further constraints applied using the neighborhood distribution distance of the corresponding $p_{i}$ points. Figure 1 illustrates the overall pipeline of our method. Below, we will describe each step of our method in detail.

\subsection{Statistical Manifold Representation of Point Clouds}
\label{sec:subsection}

Point cloud data typically originate from the surfaces of objects in three-dimensional space, and these surfaces can locally be approximated as manifolds. A manifold is a topological space, where an n-dimensional topological manifold $\mathcal M$ is a locally Euclidean, second-countable Hausdorff space. In other words, every portion of the manifold can locally be homeomorphic to Euclidean space. Therefore, we represent each local neighborhood of a point cloud as a manifold. For the selection of basis functions for fitting the surface, the fitting capability of Gaussian functions has been widely proven. Due to the excellent linear combination properties of Gaussian basis functions, they can form complex functions through weighted summation, thus capable of expressing surfaces of various shapes. Moreover, the shape of a Gaussian function is well captured by its covariance matrix. Furthermore, inspired by \cite{yu2013reconstructing}, we incorporate anisotropic kernels to accommodate the irregular distribution of particles. Additionally, anisotropic kernel regression and covariance analysis enable the accurate capture of the local distribution characteristics of fluid surfaces. Considering that a point cloud can be viewed as a sample drawn from some probability distribution, i.e., the distribution of local point cloud data can be considered as a probability distribution, the Gaussian distribution is a natural choice. It can effectively and simply describe the central tendency and the direction of spread of the data through its mean and covariance matrix. Due to the uniformity of the basis functions, we construct a unified manifold representation for the entire point cloud. Given the distributional properties of Gaussian functions, we use a statistical model to describe the structure and geometric shape of the manifold. As a result, the statistical manifold can be defined by the properties of the Gaussian functions themselves.

\begin{equation}
\Omega=\{p(\mathbf{x};\theta)|\theta=(\mu,\Sigma)\}
\end{equation}

\noindent where $p(\mathbf{x};\theta)$ is the Gaussian functions.

That is, we map the manifold representation of each neighborhood in the point cloud to a unified manifold. Since each point on the manifold represents a probability distribution, it is referred to as a statistical manifold.

\subsection{Local Gaussian Fitting}
\label{sec:subsection}

Considering the local manifold properties of point clouds, we perform a fitting representation of the local neighborhoods. Let the input point cloud be denoted as $P$. To ensure the overlap of neighborhood regions and prevent gaps during surface fitting, we first randomly select a certain number of points from the original point cloud as initial query points. This selection process ensures a uniform distribution, avoiding excessive local concentration or sparsity. Subsequently, each local patch is updated by incorporating points based on nearest-neighbor neighborhoods. Finally, the centroid of each patch is computed and designated as the final query point. For each query point $P_i$ , we construct a local coordinate system, and the projection plane is denoted as $D_{xy}$. The specific transformation formula for the coordinates on the projection plane is given below:

\begin{equation}
P_{\mathbf{x}\mathbf{y}} = R_nP
\end{equation}

\noindent where $P_{\mathbf{x}\mathbf{y}} \in D_{xy} $. The columns of matrix $R_n$ represent the directions of the x-axis, y-axis, and z-axis in the local coordinate system.

We treat the local neighborhood of point $P_i$ as a 2D manifold in 3D space. Since the manifold surface in 3D space is homeomorphic to a 2D region, for each point and its local neighborhood on the manifold, there exists a local coordinate system such that this neighborhood can be mapped to a 2D Euclidean space via a homeomorphic mapping. Therefore, a 3D local neighborhood patch can be isomorphic to the 2D parameter domain D. To eliminate the complexity caused by surface tilt in the original global coordinates, we construct a 2D parameter domain $\mathcal{D}$ for the projection plane. This means that we can establish a continuous mapping between the original local neighborhood and $\mathcal{D}$:

\begin{equation}
\phi_i{:\mathcal{P}{_i}}\to\mathcal{D}
\end{equation}

Considering the challenges of directly fitting the point cloud surface in three-dimensional space, and given that local surfaces are inherently two-dimensional, it is unnecessary to use three-dimensional Gaussian functions to describe the redundant dimension. Instead, we perform surface function fitting in a two-dimensional parameter domain and transform the parameters back to represent the original point cloud surface. In the local coordinate region, we define a series of Gaussian kernels centered at the base points.

\begin{equation}
g_{t}(u, v)=\exp \left(-\frac{1}{2}\left[\left(u-u_{t}\right),\left(v-v_{t}\right)\right] \Sigma^{-1}\left[\left(u-u_{t}\right),\left(v-v_{t}\right)\right]^{\top}\right)
\end{equation}

\noindent where $u,v \in\mathcal{D}$, $u_{t}$,$v_{t}$ represents the center of the Gaussian functions we select.

To ensure the fitting accuracy for sparse and non-uniform point clouds, we design the number of Gaussian functions $t$, based on the number of points within the neighborhood. Furthermore, drawing inspiration from the data-adaptive kernel regression method proposed in \cite{takeda2007kernel}, we capture the local characteristics of point clouds by adaptively adjusting the shape and size of the kernel function. To reduce the optimization difficulty, we fix the Gaussian function centers and only optimize the covariance matrix for each function. Through our design, the manifold of each local neighborhood in the point cloud is described as a weighted sum of multiple Gaussian functions, with the local point cloud distribution represented by the covariance matrices of the Gaussian components. Considering the statistical properties of the covariance matrix, we adopt the concept of Gaussian splatting\cite{kerbl20233d}, to ensure the positive definiteness of the covariance matrix, we decompose it into the product of a rotation matrix $R$ and a scaling matrix $S$, as expressed by the following equation:

\begin{equation}
\Sigma =RSS^{T}R^{T}  
\end{equation}

During the entire optimization process, we refine the rotation matrix $R$ and scaling matrix $S$ to maintain the positive definiteness of the covariance matrix. Finally, we fit the local neighborhood manifold of the point cloud in the 2D parametric domain, expressed as follows:

\begin{equation}
f=\sum_{t}\omega_tg_t(\mathbf{u},\mathbf{v})
\end{equation}

\noindent where $\omega_{t}$ is the weight corresponding to the Gaussian element.

In the weighted combination of Gaussian basis functions, although the basis functions themselves define part of the shape of the local surface, they do not directly define the position of the surface. The weighted Gaussian basis functions in equation (3) capture the shape features of the surface, but they do not specify the exact position of these shapes in three-dimensional space. Therefore, we design the local surface fitting for a local patch near a query point $q\in \mathbb{R}^{3} $ as follows:

\begin{equation}
f_q=q+\sum_{t}\omega_tg_t(\mathbf{u},\mathbf{v})
\end{equation}

This is our local Gaussian representation. Although our Gaussian representation inherently possesses strong approximation capabilities, extracting a reasonable Gaussian representation from information-deficient and sparse point clouds presents a significant challenge. During the optimization process of the above formulation, we leverage a generative model to learn and shape a prior. Specifically, we incorporate modules such as EdgeConv and cross-attention to account for the geometric structure of local neighborhoods, facilitating feature aggregation. The covariance matrix is optimized to model the Gaussian distribution, while the optimized Gaussian weight matrix yields the best parameter approximation, ultimately generating a Gaussian representation of the local neighborhood. Our approach ensures a refined consideration of the interactions between shallow and deep features, leveraging the powerful capabilities of neural networks to enable a more nuanced and detailed representation.

\subsection{Upsampling with Manifold Distribution Constraints}

To better generate upsampled point clouds with a uniform distribution, we first resample a sufficient number of coordinate points in the 2D parameter domain of each local neighborhood and obtain the upsampled point cloud through local Gaussian representation mapping. Additionally, current point cloud upsampling methods fail to leverage the distribution information of the point cloud to supervise the generation of upsampled points. From the statistical manifold constructed by the Gaussian representations, we propose a novel method for correcting the distribution of the point cloud.

Since each point on the manifold represents a probability distribution, to ensure that the upsampled point cloud better aligns with the true distribution, the problem can be reformulated such that the distance between corresponding points on the statistical manifold is minimized. Here, we choose the Fisher-Rao distance between two distributions. The Fisher-Rao distance between two probability distributions on the statistical manifold is the length of a geodesic, which is the local shortest distance between the two points. This geodesic is defined under the Fisher information metric between the distributions. The elements of the Fisher information matrix and the classic Fisher-Rao distance are defined as follows:

\begin{equation}
I_{ij}(\theta)=\mathbb{E}\left[\frac{\partial\log p(x;\theta)}{\partial\theta_i}\frac{\partial\log p(x;\theta)}{\partial\theta_j}\right]
\end{equation}
\begin{equation}
d_{F R}\left(\theta_{P}, \theta_{Q}\right)=\int_{\gamma} \sqrt{\mathbf{v}^{\mathrm{T}} I(\theta) \mathbf{v}} d \ell
\end{equation}

\noindent where $\gamma$ is the geodesic connecting $\theta_{P}$ and $\theta_{Q}$, $I(\theta)$ is the Fisher information matrix, $\mathbf{v}$ is the tangent vector along the geodesic at each point.

Since our local Gaussian representation consists of multiple Gaussian functions (normal distributions), the Fisher-Rao distance between two multivariate normal distributions can be expressed as:

\begin{equation}
d_{FR}(\theta_P,\theta_Q)=\sqrt{\sum_{i=1}^td_{u-FR}^2((\mu_{Pi},\sigma_{Pi}),(\mu_{Qi},\sigma_{Qi}))}
\end{equation}

\noindent where $\Sigma=\operatorname{diag}\left(\sigma_{1}^{2}, \sigma_{2}^{2}, \ldots \ldots, \sigma_{t}^{2}\right)$.

Following the approach in \cite{costa2015fisher}, we compute the Fisher-Rao distance for a univariate Gaussian probability distribution. Since the hyperbolic distance between two points $\boldsymbol{x}=\left(x_{1}, x_{2}\right)$ and $\boldsymbol{y}=\left(y_{1}, y_{2}\right)$ in the Poincaré half-plane model is given by:

\begin{equation}
d_{H}(\boldsymbol{x}, \boldsymbol{y})=\ln \left(\frac{|\boldsymbol{x}-\bar{\boldsymbol{y}}|+|\boldsymbol{x}-\boldsymbol{y}|}{|\boldsymbol{x}-\bar{\boldsymbol{y}}|-|\boldsymbol{x}-\boldsymbol{y}|}\right)
\end{equation}

\noindent where $d_H$ is the hyperbolic distance, $\bar{\boldsymbol{y}}=(y_1,-y_2)$ and $|\cdot |$ is the standard Euclidean norm.

We derive and establish the relationship between the Fisher-Rao distance for a univariate Gaussian probability distribution and the Poincaré distance $d_{H}$ within the Poincaré half-plane model in hyperbolic geometry:

\begin{equation}
d_{u-\mathrm{FR}}((\mu_1,\sigma_1),(\mu_2,\sigma_2))=\sqrt{2}d_H\left(\left(\frac{\mu_1}{\sqrt{2}},\sigma_1\right),\left(\frac{\mu_2}{\sqrt{2}},\sigma_2\right)\right)
\end{equation}

Thus, the final computation formula is given as follows:

\begin{equation}
\begin{gathered}
d_{u-\mathrm{FR}}\left(\left(\mu_{1},\sigma_{1}\right)\left(\mu_{2},\sigma_{2}\right)\right) \\
=\sqrt{2}\mathrm{log}\frac{\left|\left(\frac{\mu_1}{\sqrt{2}},\sigma_1\right)-\left(\frac{\mu_2}{\sqrt{2}},-\sigma_2\right)\right|+\left|\left(\frac{\mu_1}{\sqrt{2}},\sigma_1\right)-\left(\frac{\mu_2}{\sqrt{2}},\sigma_2\right)\right|}{\left|\left(\frac{\mu_1}{\sqrt{2}},\sigma_1\right)-\left(\frac{\mu_2}{\sqrt{2}},-\sigma_2\right)\right|-\left|\left(\frac{\mu_1}{\sqrt{2}},\sigma_1\right)-\left(\frac{\mu_2}{\sqrt{2}},\sigma_2\right)\right|}
\end{gathered}
\end{equation}

Now, to constrain the distribution of the point cloud, the objective is to minimize:
\begin{equation}
\min d_{F R}\left(\theta_{P}, \theta_{Q}\right)=\min \sqrt{\sum_{i=1}^{n} d_{u-F R}^{2}\left(\left(\mu_{P i}, \sigma_{P i}\right),\left(\mu_{Q i}, \sigma_{Q i}\right)\right)} \\
\end{equation}

\subsection{Loss Function}

In this section, we introduce the design of the loss function, which includes constraints on the quality of the point cloud, the similarity between distributions, and the smoothness of the surface shape. First, considering the importance of point cloud quality for point cloud upsampling, our method should encourage the generated points to be closer to the real point cloud. Therefore, a reconstruction loss function is needed to evaluate the similarity between the two point clouds. Here, the reconstruction loss uses the Earth Mover's distance(EMD).

\begin{equation}
\mathcal{L}_{emd}(P,P_r)=\min_{\phi:P\to P_r}\sum_{x\in P}\|x-\phi(x))\|_2
\end{equation}

Then, during the minimization process of Eq. (14), to prevent interference from outliers in some distributions, we impose a first-order differential constraint on the surface function, as shown in Eq. (16). This constraint limits the gradient variation of the surface, promoting smooth transitions in adjacent regions and enhancing surface continuity. Additionally, smoothing the gradient also contributes to improving the uniformity of the point cloud distribution.

\begin{equation}
\mathcal{L}_{\mathrm{first\_der}}=\sum_{i=1}^N\|\nabla f_q(\mathbf{x}_i,\mathbf{y}_i)\|^2
\end{equation}

\noindent where $\nabla f_q(\mathbf{x},\mathbf{y})=W_q\nabla\Phi(\mathbf{x},\mathbf{y})$, $
\nabla\Phi(\mathbf{x},\mathbf{y})=
\begin{bmatrix}
\frac{\partial\phi_1}{\partial\mathbf{x}},\ldots,\frac{\partial\phi_T}{\partial\mathbf{x}} \\
\frac{\partial\phi_1}{\partial\mathbf{y}},\ldots,\frac{\partial\phi_T}{\partial\mathbf{y}}
\end{bmatrix}$.

Finally, based on the above description, the final combined loss function is as follows:
\begin{equation}
\mathcal{L}_{total}=\mathcal{L}_{emd}+\lambda_1d_{FR}+\lambda_2\mathcal{L}_{first\_der}
\end{equation}

\section{Experiment}
\subsection{Datasets and Implementation Details}

Compared to the traditional PUGAN \cite{li2019pu}dataset, the PU1K dataset \cite{qian2021pu} is more challenging because it contains a larger volume of data and more diverse categories. This dataset includes large objects with complex shapes, as provided by PUGCN. The PU1K dataset is collected from both PUGAN and ShapeNetCore \cite{xu2019disn}. Since our method is designed for upsampling non-uniform point clouds, we processed the PU1K dataset to account for non-uniformity. During the training phase, we first sampled a dense point cloud from the original mesh as ground truth. For each patch during training, we applied non-uniform sampling as the input data.

For the testing dataset, we selected different resolutions of the PU1K dataset for evaluation. Considering the testing requirements of our method for upsampling sparse, non-uniform point clouds, we applied non-uniform processing to the original dataset as well. Additionally, to validate the generalization of our method and assess performance at different upsampling scales, we also evaluated the Sketchfab dataset \cite{qian2020pugeo} and the KITTI dataset \cite{geiger2013vision}.

All of our experiments are implemented using PyTorch. We set the cross-attention and graph convolution modules to three layers and the number of Gaussian factors per patch to $t=\left \lfloor \mathcal{N}(p_{i})/8 \right \rfloor $. We use the Adam optimizer to train our model on an RTX 3090 GPU, with a batch size of 64, an upsampling factor of 4, and for 400 epochs. The initial learning rate is set to 0.001, decaying by a factor of 0.7 every 80 iterations. Additionally, to eliminate unnecessary degrees of freedom in the input data space and reduce the learning difficulty, we normalize each point’s coordinates by its patch radius and rotate the points into a local coordinate system defined by PCA. To avoid overfitting during training, we augment the network inputs with random rotations, scaling, and Gaussian noise perturbations.

\begin{figure}[h]
  \centering
  \includegraphics[width=\linewidth]{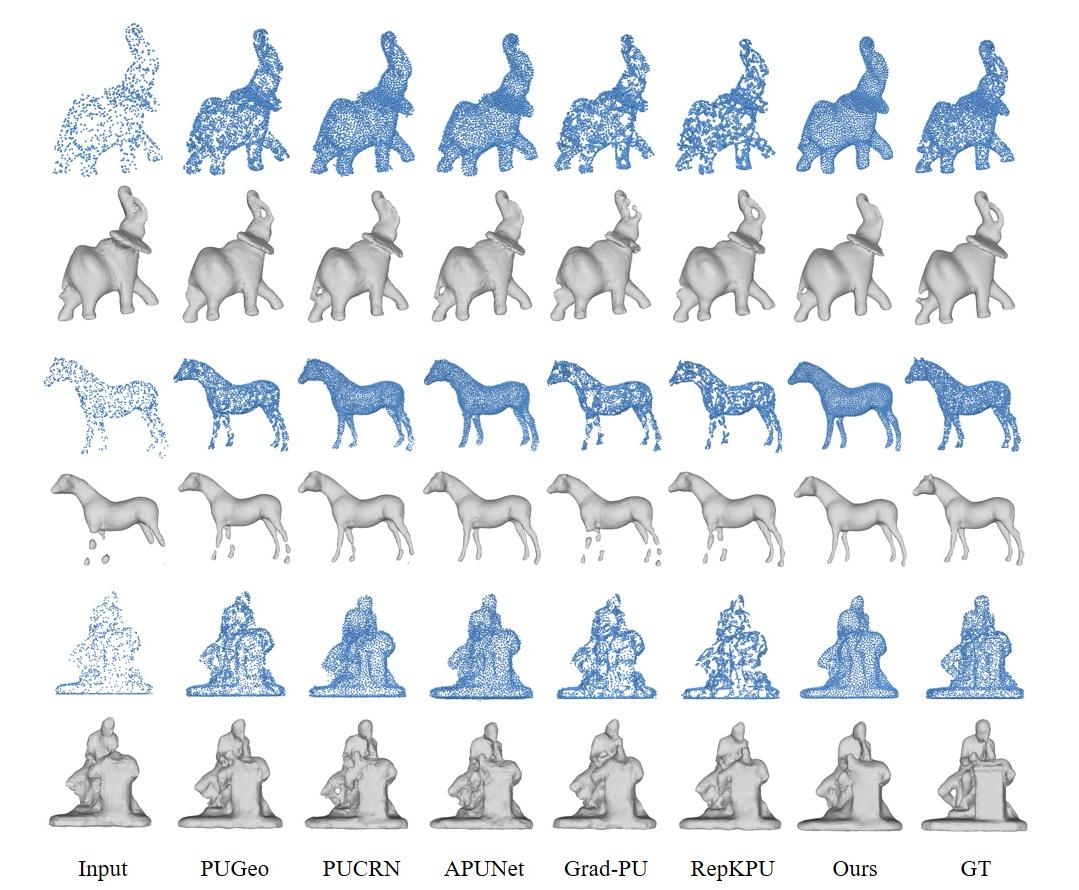}
  \caption{Visualization of the non-uniform PU1K dataset (1024) experiments based on different algorithms.}
\end{figure}

\begin{figure}[h]
  \centering
  \includegraphics[width=\linewidth]{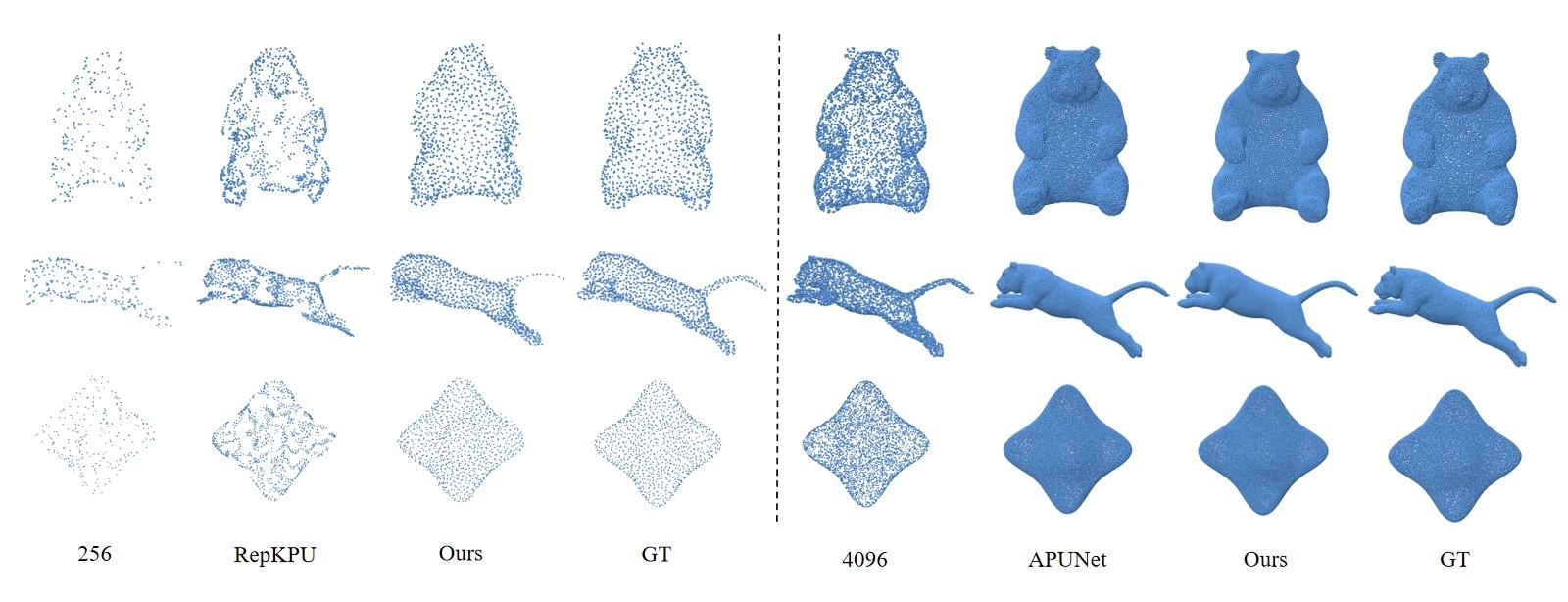}
  \caption{Visualization of upsampling results for partial data with 256 and 4096 input points.}
\end{figure}

\begin{table}[ht]
\centering
\caption{The table compares the upsampling performance of different methods on the PU1K dataset using three different input sizes. The best results are highlighted in bold.}
\begin{tabular}{c|c|ccccc}
\toprule
Input & Method & CD  & HD & P2F & JSD & UNI\\ 
\midrule
\multicolumn{1}{c|}{\textbf{256}} & PUGeo~\cite{qian2020pugeo} & 15.372 & 6.911 & 6.383 & 0.476 & 11.327\\ 
\multicolumn{1}{c|}{} & PUCRN~\cite{du2022point} & 11.076 & 6.780 & 5.761 & 0.484 & 9.347\\
\multicolumn{1}{c|}{} & APUNet~\cite{zhao2023apunet} & 9.960 & 6.143 & \textbf{5.683} & 0.375 & 7.325\\
\multicolumn{1}{c|}{} & Grad-PU~\cite{he2023grad} & 9.023 & 5.679 & 7.004 & 0.301 & 18.682\\
\multicolumn{1}{c|}{} & RepKPU~\cite{rong2024repkpu} & 8.747 & \textbf{5.443} & 6.769 & 0.287 & 17.126 \\
\multicolumn{1}{c|}{} & Ours & \textbf{8.538} & 5.910 & 6.337 & \textbf{0.218} & \textbf{5.642}\\
\midrule
\multicolumn{1}{c|}{\textbf{1024}} & PUGeo~\cite{qian2020pugeo} & 3.795 & 4.211 & 2.363 & 0.411 & 5.723\\ 
\multicolumn{1}{c|}{} & PUCRN~\cite{du2022point} & 3.382 & 4.076 & 2.727 & 0.457 & 3.697\\
\multicolumn{1}{c|}{} & APUNet~\cite{zhao2023apunet} & 2.604 & 3.726 & 2.283 & 0.302 & 3.018\\
\multicolumn{1}{c|}{} & Grad-PU~\cite{he2023grad} & 4.102 & 3.817 & 2.209 & 0.243 & 8.012\\
\multicolumn{1}{c|}{} & RepKPU~\cite{rong2024repkpu} & 3.626 & 3.762 & 2.421 & 0.211 & 7.935 \\
\multicolumn{1}{c|}{} & Ours & \textbf{2.742} & \textbf{3.540} & \textbf{2.059} & \textbf{0.163} & \textbf{2.791}\\
\midrule
\multicolumn{1}{c|}{\textbf{4096}} & PUGeo~\cite{qian2020pugeo} & 2.863 & 2.456 & 1.679 & 0.325 & 3.413\\ 
\multicolumn{1}{c|}{} & PUCRN~\cite{du2022point} &2.238 & 2.154 & 0.937 & 0.364 & 1.039\\
\multicolumn{1}{c|}{} & APUNet~\cite{zhao2023apunet} & 1.396 & 1.496 & 0.364 & 0.213 & 0.591\\
\multicolumn{1}{c|}{} & Grad-PU~\cite{he2023grad} & 2.963 & 1.539 & 0.367 & 0.175 & 4.537\\
\multicolumn{1}{c|}{} & RepKPU~\cite{rong2024repkpu} & 2.723 & 1.442 & 0.232 & 0.134 &4.261 \\
\multicolumn{1}{c|}{} & Ours & \textbf{1.014} & \textbf{1.238} & \textbf{0.038} & \textbf{0.097} & \textbf{0.128}\\
\bottomrule
\end{tabular}

\end{table}

\begin{table}
  \caption{Floating point operations, training and inference time comparison.}
  \label{tab:freq}
  \begin{tabular}{c|c|c|c}
    \toprule
    Method & FLOPs(G) & Training(h) & Inference(s) \\
    \midrule
    PUGeo~\cite{qian2020pugeo} & 8.782 & 2.1 & 0.325 \\
    PUCRN~\cite{du2022point} & 4.315 & 5.8 & 0.278 \\
    APUNet~\cite{zhao2023apunet} & 5.713 & 6.3 & 0.536 \\
    Grad-PU~\cite{he2023grad}& 2.718 & 4.8 & 0.269 \\
    RepKPU~\cite{rong2024repkpu}& \textbf{2.062} & 4.3 & \textbf{0.187} \\
    Ours & 3.207 & \textbf{1.3} & 0.201 \\
    \bottomrule
  \end{tabular}
\end{table}

\subsection{Evaluation Metrics and Comparisons}

Similar to recent point cloud upsampling works, for quantitative evaluation, we use three common metrics: Chamfer Distance ($\text{CD} \times  10^{-3}$), Hausdorff Distance ($\text{HD} \times  10^{-4}$), and Point-to-Surface Distance ($\text{P2F} \times  10^{-3}$). These metrics reflect the reconstruction quality of the upsampled point cloud. Additionally, since our method considers the distribution similarity of each patch, we include the Jensen-Shannon Divergence ($\text{JSD} \times  10^{-3}$) metric, which measures the similarity between point cloud distributions. Finally, we introduce the Uniformity ($\text{UNI} \times  10^{-2}$) metric proposed in PU-GAN \cite{li2019pu}, which reflects the uniformity of the point cloud. For all metrics, smaller values indicate better performance.

We compare the method proposed in this paper with five existing point cloud upsampling methods, including PUGeo \cite{qian2020pugeo}, PUCRN \cite{du2022point}, APUNet\cite{zhao2023apunet}, Grad-PU\cite{he2023grad}, and RepKPU\cite{rong2024repkpu}. For a fair comparison, we retrained these methods on our dataset in the same experimental environment, using the official code and recommended settings.

\subsection{Results on synthetic dataset}

The quantitative comparison results are shown in Table 1. We performed a quantitative analysis on sparse, non-uniform point clouds with 256, 1024, and 4096 input points. From the table, we can see that our method achieves superior results in upsampling sparse, non-uniform point clouds compared to most other methods. Although, for 256 input points, some metrics like HD and P2M are not as good as RepKPU, the visual results in Figure 3 show that RepKPU produces highly uneven results. This is because when dealing with extremely sparse point clouds, anomalies can occur during point generation, affecting some metrics significantly. In contrast, other methods tend to generate highly clustered points, which may yield better results for some metrics. However, the visual results clearly demonstrate that our method generates points that most closely align with the original features.

\begin{table}[t]
\centering
\caption{Quantitative results of upsampling at different noise levels. The best results are highlighted in bold.}
\begin{tabular}{c|c|ccccc}
\toprule
Noise Level & Method & CD  & HD & P2F & JSD & UNI\\ 

\midrule
\multicolumn{1}{c|}{\textbf{0.5\%}} & PUGeo~\cite{qian2020pugeo} & 2.961 & 2.447 & 1.702 & 0.412 & 3.437\\ 
\multicolumn{1}{c|}{} & PUCRN~\cite{du2022point} & 2.269 & 2.170 & 1.003 & 0.395 & 1.219\\
\multicolumn{1}{c|}{} & APUNet~\cite{zhao2023apunet} & 1.418 & 1.489 & 0.387 & 0.276 & 0.613\\
\multicolumn{1}{c|}{} & Grad-PU~\cite{he2023grad} & 2.901 & 1.543 & 0.397 & 0.231 & 4.576\\
\multicolumn{1}{c|}{} & RepKPU~\cite{rong2024repkpu} & 2.736 & 1.441 & 0.242 & 0.215 & 4.432 \\
\multicolumn{1}{c|}{} & Ours & \textbf{1.009} & \textbf{1.245} & \textbf{0.039} & \textbf{0.103} & \textbf{0.125}\\
\midrule
\multicolumn{1}{c|}{\textbf{1\%}} & PUGeo~\cite{qian2020pugeo} & 3.107 & 2.491 & 1.875 & 0.435 & 3.764\\ 
\multicolumn{1}{c|}{} & PUCRN~\cite{du2022point} & 2.364 & 2.202 & 1.127 & 0.412 & 1.493\\
\multicolumn{1}{c|}{} & APUNet~\cite{zhao2023apunet} & 1.739 & 1.532 & 0.291 & 0.298 & 0.836\\
\multicolumn{1}{c|}{} & Grad-PU~\cite{he2023grad} & 3.046 & 1.594 & 0.406 & 0.265 & 4.731\\
\multicolumn{1}{c|}{} & RepKPU~\cite{rong2024repkpu} & 2.938 & 1.479 & 0.325 & 0.251 & 4.682 \\
\multicolumn{1}{c|}{} & Ours & \textbf{1.225} & \textbf{1.301} & \textbf{0.049} & \textbf{0.127} & \textbf{0.135}\\
\midrule
\multicolumn{1}{c|}{\textbf{2\%}} & PUGeo~\cite{qian2020pugeo} & 3.208 & 2.553 & 1.938 & 0.467 & 4.024\\ 
\multicolumn{1}{c|}{} & PUCRN~\cite{du2022point} & 2.567 & 2.239 & 1.183 & 0.422 & 1.992\\
\multicolumn{1}{c|}{} & APUNet~\cite{zhao2023apunet} & 1.962 & 1.633 & 0.579 & 0.310 & 1.267\\
\multicolumn{1}{c|}{} & Grad-PU~\cite{he2023grad} & 3.298 & 1.607 & 0.711 & 0.288 & 6.931\\
\multicolumn{1}{c|}{} & RepKPU~\cite{rong2024repkpu} & 3.125 & 1.532 & 0.653 & 0.267 & 6.637 \\
\multicolumn{1}{c|}{} & Ours & \textbf{1.312} & \textbf{1.332} & \textbf{0.067} & \textbf{0.145} & \textbf{0.147}\\
\bottomrule
\end{tabular}
\label{tab:comparison}
\end{table}

In addition to the quantitative results, we also present point cloud upsampling and surface reconstruction results for some models in Figure 2. Here, we use Poisson reconstruction \cite{kazhdan2013screened}. As seen in the figure, when dealing with complex data, such as the elephant, our method produces smoother and more uniformly distributed points compared to other methods, providing excellent guidance for subsequent reconstruction. Our upsampling results effectively fill in gaps, while other methods tend to generate more noise and uneven point sets.

Additionally, to evaluate runtime performance, we compare these methods in terms of floating point operations (FLOPs), as well as the training and the inference time. As shown in Table 2, our method exhibits the shortest training time while also achieving competitive performance in FLOPs and inference efficiency.

\begin{figure}[h]
  \centering
  \includegraphics[width=\linewidth]{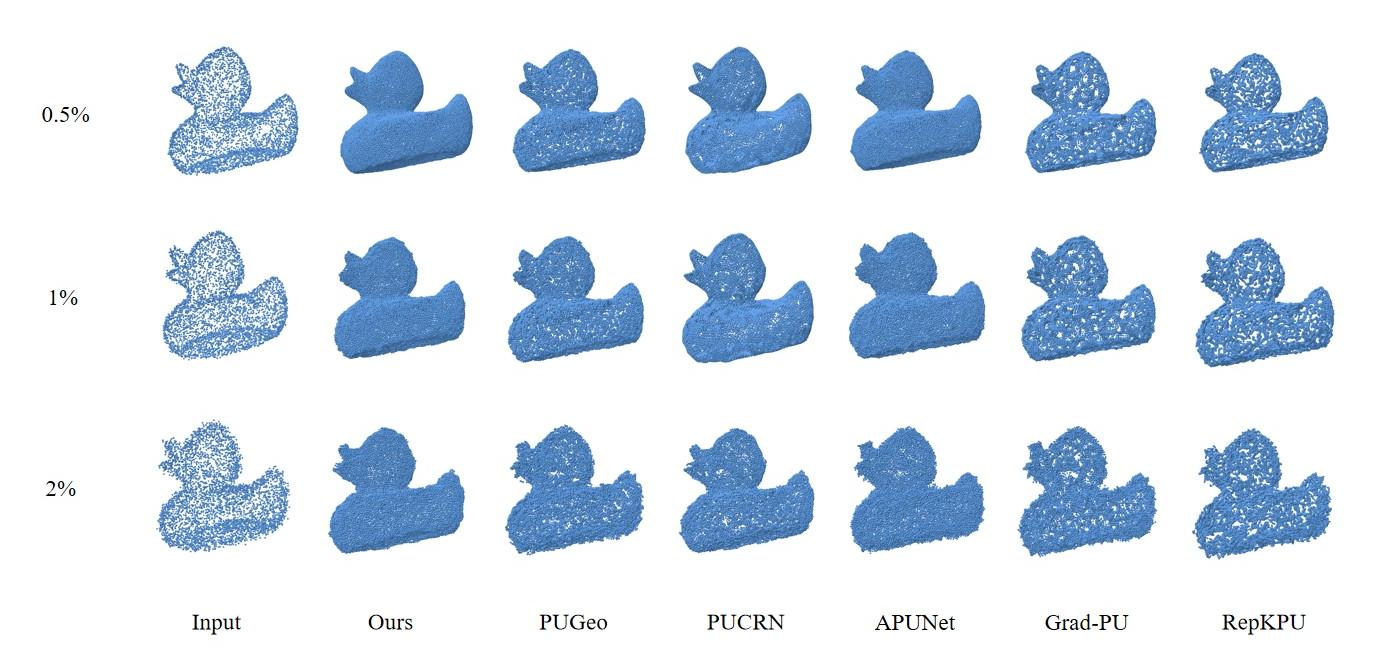}
  \caption{Upsampling results for input point clouds with noise levels of 0.5\%, 1\%, and 2\%.}
\end{figure}

\begin{figure}[h]
  \centering
  \includegraphics[width=\linewidth]{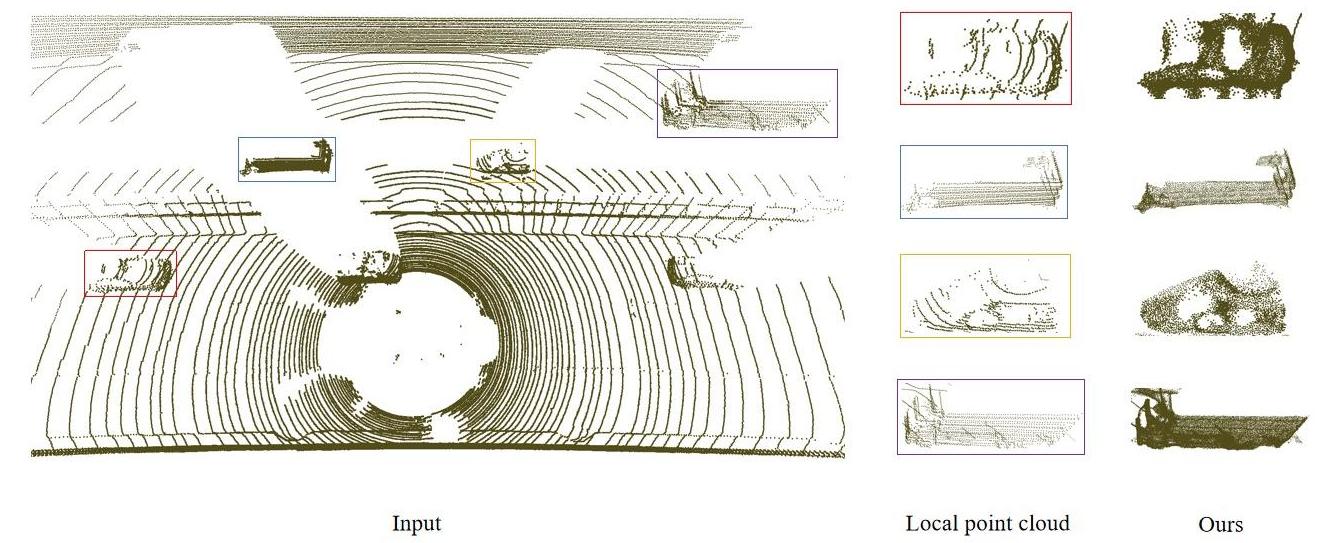}
  \caption{ Upsampling results for LiDAR real-time scanned point cloud data.}
\end{figure}

\begin{table}
  \caption{Ablation study of different modules of our method on the PU1K dataset. I is the absence of a Gaussian representation, II is without the constraints of manifold distribution, III is without the first-order derivative loss.}
  \label{tab:freq}
  \begin{tabular}{c|ccccc}
    \toprule
    Ablation & CD & HD & P2F & JSD & UNI\\
    \midrule
    I & 1.022 & 1.240 & 0.041 & 0.103 & 0.132\\
    II & 1.016 & 1.239 & 0.042 & 0.108 & 0.130\\
    III & 1.025 & 1.241 & 0.044 & 0.099 & 0.130\\
    Ours & \textbf{1.014} & \textbf{1.238} & \textbf{0.038} & \textbf{0.097} & \textbf{0.128}\\
  \bottomrule
\end{tabular}
\end{table}

\begin{figure}[htbp]
    \centering
    \begin{subfigure}[b]{0.3\textwidth}
        \includegraphics[width=\textwidth]{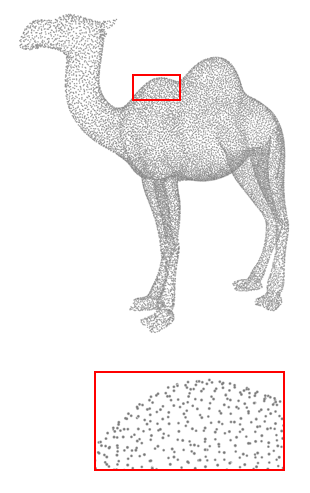}  % 替换为你的图片路径
        \caption{}
        \label{fig:subfig1}
    \end{subfigure}
    \hfill
    \begin{subfigure}[b]{0.3\textwidth}
        \includegraphics[width=\textwidth]{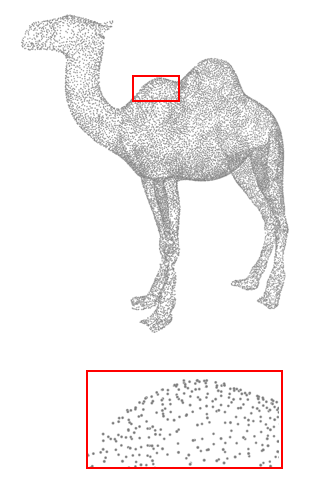}  % 替换为你的图片路径
        \caption{}
        \label{fig:subfig2}
    \end{subfigure}
    \hfill
    \begin{subfigure}[b]{0.3\textwidth}
        \includegraphics[width=\textwidth]{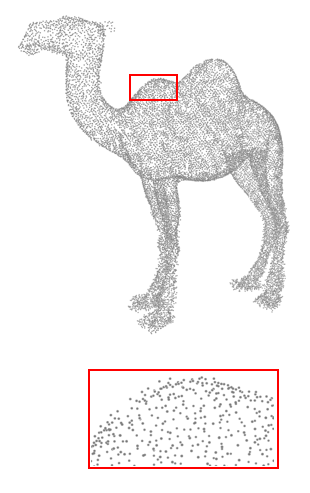}  % 替换为你的图片路径
        \caption{}
        \label{fig:subfig3}
    \end{subfigure}
    \caption{Visualization results after removing different modules. (a) Ours. (b) Without manifold distribution constraints. (c) Without the first-order derivative loss.}
    \label{fig:subfigures}
\end{figure}

\subsection{Results on noise dataset} 

To validate the robustness of our method against noisy results, we performed upsampling on point clouds with noise levels of 0.5\%, 1\%, and 2\%. As shown in Table 2, our method provides the best results across different noise levels. From Figure 4, we can observe that our method effectively constrains the noisy outliers to be closer to the surface, and our method also achieves the best performance in terms of uniformity.

\subsection{Results on real-scanned point clouds}

To validate the effectiveness of our method in real-world scenarios, we conducted experiments on the KITTI dataset. Since there is no ground truth point cloud, we provide a visual comparison in Figure 5. As seen, even when facing real scanned point clouds, our method is still able to fill in gaps and output a more uniform point distribution.

\subsection{Ablation Study}

To verify the effectiveness of each module, we conduct an ablation study to demonstrate how each component impacts the final results. Specifically, we use the PU1K dataset as a benchmark. We focus on the selection of Gaussian functions(I), the distribution constraints on the manifold(II), and the design of the first-order derivative loss(III). For the function selection, we adopt the quadratic term representation of surfaces provided in GeoUDF \cite{ren2023geoudf}. The quantitative results of the ablation study are shown in Table 3. It can be observed that each component we designed significantly improves the experimental performance. From Figure 6, it is evident that without the manifold distribution constraint, the points become more clustered. The absence of the first-order derivative loss results in interference from anomalous points. The experimental results strongly support the effectiveness and necessity of each strategy we employed.

\section{Conclusion}

We take an innovative approach by starting from the perspective of statistical manifolds. In the local coordinate system of point cloud neighborhoods, we construct Gaussian functions, and through continuous optimization of function parameters via a neural network, we fit local surfaces. The parameters of each local function are treated as a point on the statistical manifold, and point cloud upsampling is performed through distribution constraints on the manifold. Our method effectively learns from sparse and non-uniform datasets and generates more uniformly distributed points than existing methods. Extensive experiments have validated the strong representational power, robustness, and generalization ability of our approach, significantly improving the efficiency of subsequent reconstruction tasks.

\begin{acks}
This work was supported in part by Beijing Municipal Science and Technology Commission and Zhongguancun Science Park Management Committee under Grant Z221100002722020, in part by the National Nature Science Foundation of China under Grant 62072045, in part by the National Nature Science Foundation of Beijing under Grant 7242167.
\end{acks}

\bibliographystyle{ACM-Reference-Format}
\bibliography{main}


\end{document}